\documentclass{article} % For LaTeX2e
\usepackage{iclr2027_conference,times}

\usepackage{amsmath,amsfonts,bm}

\def\eqref#1{equation~\ref{#1}}
\def\1{\bm{1}}

\DeclareMathAlphabet{\mathsfit}{\encodingdefault}{\sfdefault}{m}{sl}
\SetMathAlphabet{\mathsfit}{bold}{\encodingdefault}{\sfdefault}{bx}{n}

\usepackage{hyperref}
\usepackage{url}
\usepackage{graphicx}
\usepackage{amssymb}
\usepackage{booktabs}
\usepackage{multirow}
\usepackage{xcolor}
\usepackage{colortbl}
\usepackage{float}
\usepackage{xspace}

\usepackage{enumitem}

\iclrfinalcopy

\title{HairFlow: Modeling 3D Hair Strands with Multi-Modal Conditioned Diffusion}

\author{
Yating Wang$^{1,2}$,
Kui Wu$^{2}$,
Ran Yi$^{1}$,
Zixuan Lu$^{2}$,
\AND
Tongtong Wang$^{2}$,
Zhongtian Zheng$^{2}$,
Lizhuang Ma$^{1}$\thanks{Corresponding author},
Yuefan Shen$^{2}$
\\
$^{1}$Shanghai Jiao Tong University
\quad
$^{2}$LIGHTSPEED
}

\begin{document}

\maketitle

\begin{figure}[H]
    \centering
    \includegraphics[width=\linewidth]{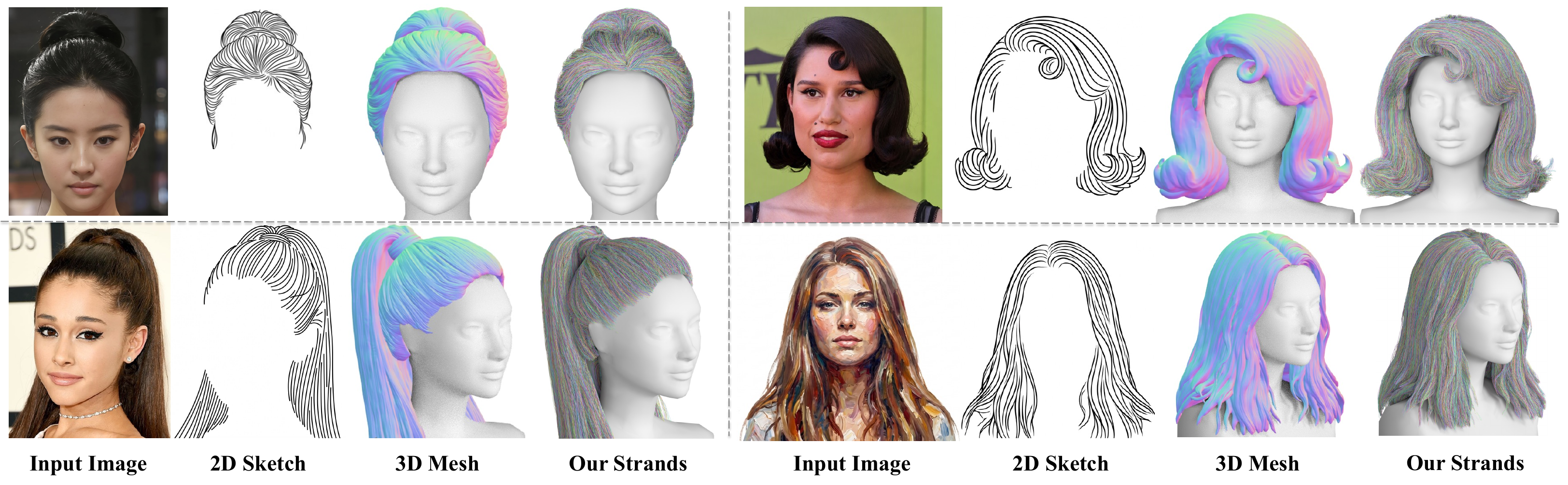}
    \vspace{-1em}
    \caption{HairFlow uses conditioned diffusion to recover 3D strands from image inputs, guided by 3D mesh and sketches inferred by foundation models. It handles complex hairstyles and stylized images while closely adhering to the input conditions.}
    \label{fig:teaser}
\end{figure}

\begin{abstract}
%发丝级三维头发建模面临发型结构复杂、训练数据有限以及输入条件稀疏或含有噪声等挑战。近年来，大规模预训练模型在视觉理解与三维生成方面展现出强大的通用能力，为三维头发建模带来了新的可能。近期方法利用大规模三维重建模型生成的粗糙头发网格作为结构先验，并结合二维方向预测结果，通过逐实例优化恢复用于发丝提取的完整三维方向场。尽管这一方式缓解了建模歧义，但逐实例优化效率较低，而二维方向预测模型受限于有限的头发训练数据，难以泛化到复杂多样的发型。 我们提出 HairFlow，一个用于高效、可控三维头发建模的多模态条件扩散框架。HairFlow 采用两种互补的几何条件：粗糙头发网格约束整体形状，多视角 sketch 描述发丝走向和局部结构。来自单张图像、文本或艺术家指引的信息均可转换为这种 mesh–sketch 条件形式，使 HairFlow 能够处理不同形式的输入。给定网格和 sketch，HairFlow 通过潜空间扩散直接生成三维方向场的潜表示，并将其解码为用于发丝提取的完整方向场，从而替代逐实例潜变量优化，并支持更高容量的方向场表示。为解决实际 sketch 与三维发丝之间缺少成对训练数据的问题，我们进一步提出一套自动化的数据合成方法，从真实三维发丝中生成几何对齐、且接近测试输入形式的多视角 sketch。实验表明，HairFlow 提高了复杂发型的方向场与发丝建模精度，并支持不同形式的输入以及对头发整体形状和局部流向的灵活控制

Modeling 3D hair strands is challenging due to complex hairstyle geometry, scarce high-quality training data, and incomplete or noisy input observations. 
Foundation models offer broader hairstyle priors, but their outputs are not directly usable for strand-based rendering and simulation.
Existing approaches require costly per-instance optimization to fit hair priors to meshes from foundation models and 2D orientations from specialized predictors trained on limited hair data, restricting efficiency and generalization.
We present HairFlow, a multimodal conditioned diffusion framework for efficient and controllable hair strands modeling. HairFlow conditions strand generation on two complementary representations: a coarse mesh that constrains global shape and multi-view sketches that specify local strand flow. 
HairFlow combines an Orientation VAE with a conditioned diffusion transformer to generate volumetric fields that are traced into explicit strands.
To overcome the lack of paired sketch–strand training data, we introduce an automatic synthesis pipeline that generates geometrically aligned multi-view sketches with diverse styles representative of inference-time inputs. Experiments show that HairFlow generates plausible 3D strands that faithfully follow the input conditions across diverse hairstyles, substantially reduces inference time, and supports flexible multimodal control.

\end{abstract}

\section{Introduction}

3D hair modeling is essential for digital avatars, games, and visual effects. Strand-based representations are standard in high-end production, enabling high-quality rendering and physically based animation. A hairstyle can contain over 100,000 strands with complex structures such as curls, locks, and braids, making manual authoring through guide placement and grooming highly labor-intensive.

Recent works~\cite{zheng2023hairstep, zhou2024groomcap, rosu2022neural, kuang2022deepmvshair,sklyarova2025im2haircut, rosu2025difflocks} have explored reconstructing strand-level geometry from single- or multi-view images. However, hair reconstruction is inherently underconstrained: images provide incomplete observations of the underlying 3D strand structure, particularly in occluded regions and beneath overlying hair. Existing learning-based methods therefore rely heavily on hairstyle priors learned from 3D hair datasets, whose limited scale and diversity restrict generalization to complex, in-the-wild hairstyles.
Foundation models offer an alternative by providing rich geometric and visual priors learned at scale. Large 3D reconstruction models (LRMs), for example, can recover coarse hair geometry, while image generation models can provide visual cues about local hair flow. Recent methods have begun leveraging these representations for strand reconstruction~\cite{shen2026hairlrm,hairlrm2_conghui}. HairLRM~\cite{shen2026hairlrm} fits a learned hair prior to a reconstructed hair mesh and 2D orientation maps to obtain a volumetric orientation field for strand extraction. However, its per-instance latent optimization is costly and limits representation capacity, while its orientation estimator, trained on limited hair data, generalizes poorly to diverse hairstyles.

We present HairFlow, a multimodal conditioned diffusion framework for 3D hair strand modeling. HairFlow uses a coarse hair mesh and multi-view sketches as a unified interface: the mesh constrains global shape and spatial extent, while the sketches specify local strand flow. These conditions can be generated from images or text using foundation models or provided directly by artists, enabling a flexible interface for strand modeling from diverse inputs. At its core, HairFlow learns a compact latent representation of volumetric hair orientations using an Orientation VAE and generates these latents with a conditional diffusion transformer trained via flow matching. Mesh- and camera-aware sketch features are incorporated through cross-attention, enabling direct generation without per-instance optimization. We then decode the generated orientation field into explicit 3D strands. To address the scarcity of paired training data, we also introduce an automatic pipeline to synthesize geometrically aligned multi-view sketches with diverse stroke styles from existing 3D strand assets.

Qualitative comparisons on real-world images and quantitative evaluations on synthetic image–strand pairs demonstrate that HairFlow generates 3D strand geometry more closely aligned with the inputs while reducing end-to-end inference time. We further demonstrate its support for multi-view sketches, diverse sketch and image styles, and artist-provided guidance, as well as the downstream usability of the generated strands in physical simulation.

\section{Related Work}

\paragraph{Hair Representations}
Mesh-based~\citep{yuksel2009hair} and card-based representations~\citep{zheng2025auto, tojo2025strands2cards} are compact and support efficient real-time rendering~\citep{wu2016real, huang2025real}, but their low-polygon nature precludes accurate simulation. NeRF~\citep{mildenhall2021nerf} and Gaussian splatting~\citep{kerbl20233d} capture hair appearance effectively for view synthesis~\citep{wang2025mega, wang20253d, zakharov2024human, wang2026dgh}, yet are unsuitable for physically based rendering~\citep{marschner2003light}. Strand-based representations therefore remain the standard for high-end film and game production.
Since raw strands are high-dimensional and unstructured, most methods adopt more structured intermediate representations: 3D orientation fields supporting strand growing~\citep{shen2026hairlrm, shen2023ct2hair, wu2022neuralhdhair}, scalp-UV parameterizations~\citep{rosu2025difflocks, zhou2023groomgen}, sparse guide hairs for interpolation~\citep{hsu2024real}.
For image-based strand modeling, 2D geometric cues bridge image appearance and 3D hair representations. \citet{2012Single} extract orientation maps using Gabor filters, while \citet{zheng2023hairstep} learn to predict strand direction and depth maps. However, filtering is sensitive to image noise and illumination, while learned predictors rely on limited synthetic hair data. We instead use multi-view sparse sketches as higher-level descriptions of hair flow, paired with a coarse mesh specifying the hair envelope, to condition a volumetric orientation field for strand generation.

\paragraph{Hair Strand Modeling}
High-fidelity strand reconstruction can be achieved using multi-view imagery~\citep{nam2019strand, zhou2024groomcap, rosu2022neural, kuang2022deepmvshair} or CT capture~\citep{shen2023ct2hair}, but requires specialized acquisition setups. Single-image reconstruction~\cite{sklyarova2025im2haircut, rosu2025difflocks} offers a more accessible alternative, yet a single view reveals only part of a hairstyle, leaving its depth and occluded structures ambiguous. Early approaches~\cite{hu2014robust} address this ambiguity by retrieving similar hairstyles from strand assets database.
Learning-based methods instead construct and augment 3D strand datasets to learn priors over hair geometry~\citep{rosu2025difflocks, he2025perm, zheng2024towards}. Different architectures have been explored: HairStep~\citep{zheng2023hairstep} uses predicted 2D orientation maps as an intermediate representation for recovering 3D hair geometry; DiffLocks~\citep{rosu2025difflocks} formulates strand generation through diffusion; and HairGPT~\citep{luo2026hairgpt} and CHARM~\citep{he2025charm} adopt autoregressive modeling. Despite these advances, the limited coverage of training data remains a bottleneck for generalization to diverse real-world hairstyles.

Recent work leverages large pretrained models to provide additional geometry evidence for sparse view reconstruction, for example, HairOrbit~\citep{jin2026hairorbit} uses a video generation model to synthesize multiple views.
Our method follows this line of work and is most closely related to HairLRM~\citep{shen2026hairlrm}, which takes a 3D mesh from large reconstruction model and a 2D orientation map predicted by HairStep~\citep{zheng2023hairstep} as input. 
HairLRM jointly represents the hair shell as a signed distance field and the internal strand flow as a volumetric orientation field in one VAE latent space. At inference, it optimizes the latent code to match the decoded distance field to the mesh and the decoded orientations to predicted 2D orientation maps lifted onto the hair mesh, then extracts strands from the completed field.
This per-instance optimization is costly, while its flow guidance remains dependent on a predictor trained on limited hair data.
A more recent work~\citep{hairlrm2_conghui} uses a training-free pipeline to convert multi-view renderings of an hair mesh into dense orientation maps via an image-editing model, back-project them onto the mesh, and propagate surface orientations inward using hand-crafted rules.
This approach relies on slow iterative propagation, and cues derived from mesh renderings may deviate from the input image.

Beyond real images, hair can be generated from text~\citep{sklyarova2024text, sun2025strandhead}, sketches~\citep{shen2020deepsketchhair, zhang2025stranddesigner, long2025tangled}, or several modalities jointly~\citep{luo2026hairgpt, wei2022hairclip}, and recent work targets stylized portraits~\citep{he2025charm}. 

% Sketch-conditioned methods here operate in 2D image space or on a scalp parameterization, whereas we condition a volumetric orientation field on multiple posed sketches at once, making the sketch constraint metrically consistent with the mesh constraint.

% \paragraph{Controllable generation}
% ControlNet~\citep{zhang2023adding} injects spatially aligned signals such as edges or depth into generation, while IP-Adapter~\citep{ye2023ip} supplies higher-level guidance from reference images, both addressing the sparsity of natural-language control. Similar ideas extend to 3D generation~\citep{chen2023control3d, zhang2024clay, cheng2023sdfusion}, where recent models accept structured multi-modal conditions~\citep{hunyuan3d2025hunyuan3d} such as point clouds or voxel occupancy, and several works explore sketch-guided generation and editing~\citep{zheng2024sketch3d, liu2024sketchdream}. Our setting differs in that both conditions are geometric yet live at different scales and in different domains, one volumetric and one projected. Rather than concatenating them into a single control signal, we keep them as separate streams and drop them independently during training, yielding decoupled classifier-free guidance and hence separate control over global shape and local flow.

\section{Method}
\label{sec:method}

\begin{figure}[t]
    \centering
    \includegraphics[width=\linewidth]{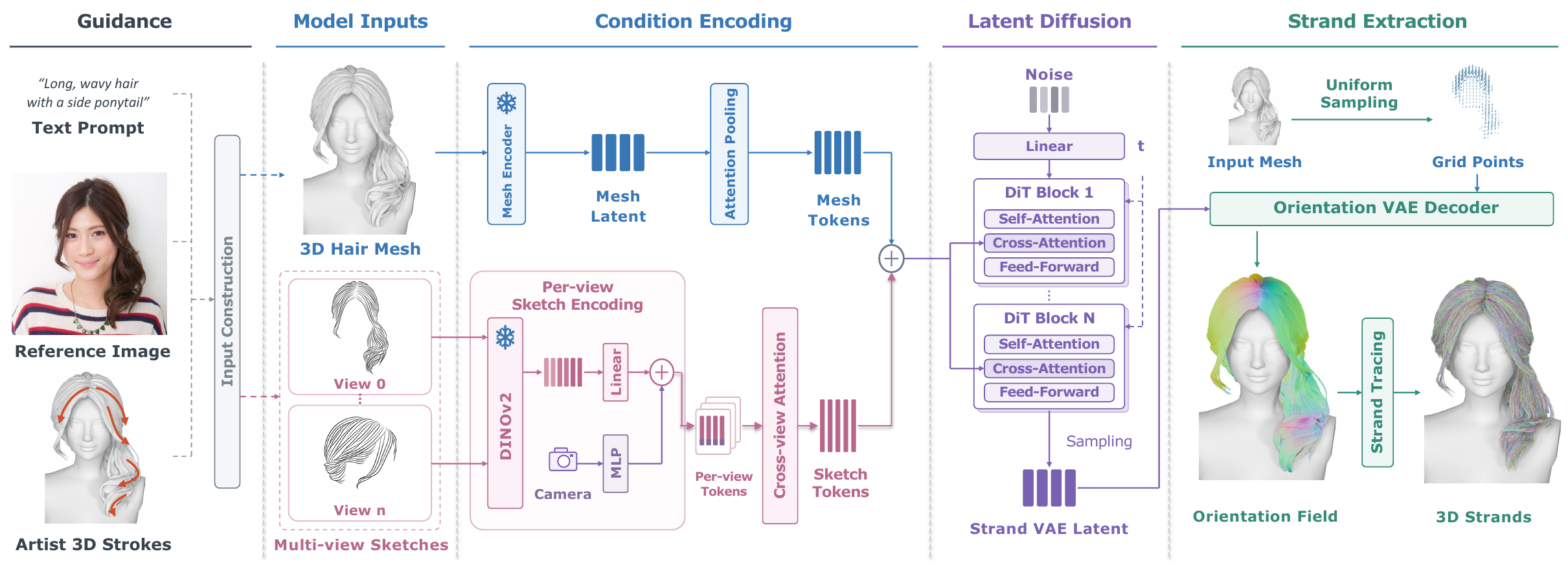}
    \vspace{-1em}
    \caption{Overview of HairFlow. HairFlow takes two geometric inputs—a coarse hair mesh and multi-view sketches—to specify global shape and local strand flow. These conditions can be obtained from images or text using foundation models, or provided by artists. Mesh and camera-aware sketch encoders produce tokens that condition a diffusion transformer through cross-attention. The transformer generates latents in the space of a pretrained Orientation VAE, which decodes them into an orientation field for explicit strand extraction.
}
    \vspace{-0.5em}
    \label{fig:pipeline}
\end{figure}

Given a hair mesh and sketches, HairFlow generates 3D hair strands that follow the mesh’s global shape and the sketches’ local flow (\autoref{fig:pipeline}). 
We represent hair strands by volumetric orientation fields that specifies local strand directions, which can be extracted to explicit strands.
We learn a compact latent space with an Orientation VAE (\autoref{sec:VAE}) and generate latents using a conditional diffusion transformer (\autoref{sec:dit}). Mesh and camera-aware sketch tokens (\autoref{sec:cond}) guide generation through cross-attention. The generated latents are decoded into an orientation field and traced into explicit strands. \autoref{sec:train} details data synthesis, training and inference process.

% Given a coarse hair mesh and one or more hair sketches, we generate a full 3D, strand-level hairstyle consistent with both the global volumetric shape specified by the mesh and the local directional cues indicated by the sketches. \autoref{fig:pipeline} shows an overview of the resulting framework, \ours. We achieve this by learning a compact latent space with a hair orientation variational autoencoder (VAE, \autoref{sec:VAE}) and performing conditional generation in that space with a diffusion transformer (DiT). A mesh encoder extracts global geometric tokens from the coarse hair mesh, and a sketch encoder produces camera-aware tokens from the input sketches (\autoref{sec:cond}); these conditioning tokens are fused and injected into the diffusion backbone via cross-attention to guide latent denoising (\autoref{sec:dit}). The orientation VAE decoder decodes the denoised latent into a dense orientation field, from which it extracts hair strands. \autoref{sec:train} describes training and inference, including training data synthesis, training objectives, and how we obtain the two conditions at inference time.

\subsection{Orientation VAE}
\label{sec:VAE}

% \wyt{delete figure2 which describes orientation VAE architecture, since it's similar to hairlrm}
% \begin{figure}[t]
%     \centering
%     \figplaceholder{4.4cm}{orientation VAE architecture, drawn as a side-by-side contrast.
%     \textbf{Top (HairLRM):} uniform + salient point clouds $\rightarrow$ encoder $\rightarrow$
%     $2\times512$ latent $\rightarrow$ two decoder heads (distance field, orientation field) plus a
%     hair/bust classification head; mark the latent with "recovered by per-instance optimization".
%     \textbf{Bottom (ours):} same point-cloud input $\rightarrow$ encoder $\rightarrow$
%  $2048\times64$ latent $\rightarrow$ a single orientation head; mark the removed distance-field
%     and bust branches with grey strike-through, and mark the latent with "sampled by diffusion".
%     Annotate the query-point input $\mathbf{x}$ on the decoder side.}
%     \caption{Our strand autoencoder against that of HairLRM~\citep{shen2026hairlrm}. Because the
%     coarse hair mesh already specifies where hair exists, we drop the distance-field and bust
%     branches and spend the entire latent budget on the 3D orientation field, which lets us enlarge
%     the latent from $2\times512$ to $2048\times64$.}
%     \label{fig:VAE}
% \end{figure}

We build our Orientation VAE on the 3DShape2VecSet-based architecture~\citep{zhang20233dshape2vecset} adopted by HairLRM~\citep{shen2026hairlrm}. HairLRM jointly models a hair SDF and a volumetric orientation field, optimizing latent codes per instance to fit the input mesh and 2D orientations predicted by HairStep. In contrast, we generate latent codes directly through multimodal conditional diffusion, without per-instance fitting. Our VAE therefore models only the orientation field, omitting the SDF required for mesh fitting in HairLRM and dedicating more model capacity to local strand directions.
More importantly, HairLRM employs a compact $512\times2$ latent representation to keep per-instance optimization tractable, whereas our generative formulation removes this constraint. We therefore increase the latent capacity to $2048\times64$, enabling finer-grained 3D hair orientation.

Given the latent tensor $\mathbf{z}$ and a query point $\mathbf{x}\in\mathbb{R}^{3}$, the decoder predicts the local hair 3D orientation
\begin{equation}
\label{eq:decoder}
\mathbf{o}(\mathbf{x}) = D_{\mathrm{hair}}(\mathbf{z}, \mathbf{x}), \qquad
\mathbf{o}(\mathbf{x}) \in \mathbb{R}^{3}.
\end{equation}
Because the diffusion model operates directly in the strand latent space, only the decoder is required at inference. Given a sampled latent $\mathbf{z}$ and the coarse hair mesh $M$, we uniformly sample 3D query points $X=\{\mathbf{x}_i\}_{i=1}^{N}$ inside $M$ and evaluate $D_{\mathrm{hair}}$ at these points to obtain a dense volumetric orientation field, from which explicit hair strands can be extracted. 

\subsection{Multi-Modal Conditioning}
\label{sec:cond}

%网格提供了overall的几何约束，sketch提供更细节的方向描述，3d网格跟sketch图像是相差比较多的两种模态，如何align这两种模态，让diffusion生成的结果同时符合两种约束条件，不是那么trival的；在推理阶段我们的sketch是图像编辑大模型生成的，也可能是美术手工绘制的，sketch和sketch之间可能存在很大的差异，为了保持鲁棒，我们用了预训练的dino特征代替sketch图像作为网络的输入；sketch是黑色背景和白色线条构成的简单图像，而dino是在真实图像上训练出来的，所以存在一定的domain gap，我们观察到dino的底层会更关注局部的结构特征，深层会更关注高层的语义特征，为了让输入信息更丰富，我们选择了dino的多层feature一起作为输入；为了支持不同视角的输入，我们在sketch encoder部分注入了相机参数，我们用了正交投影模型和6维plucker ray来描述每张sketch的视角,(说一下plucker ray的好处)，为了防止6维的相机被dino特征淹没，我们对dino特征做了降维投影，plucker做了升维，然后拼在一起；最后是把输入的所有视角的sketch的dino特征和相机得到的token都放在一起，经过一串attention，学习不同视角。

Our method adopts a unified condition interface consisting of a coarse hair mesh and multi-view sketches. The mesh constrains hair length, volume, and silhouette, while the sketches specify local strand directions and finer styling details. These modalities differ substantially: the mesh provides dense 3D geometry without directions, whereas sketches convey sparse directional cues in 2D projections.
Other modalities, such as images, text, and videos, can also be supported by converting them into these two basic inputs.
Our conditioning module aligns these complementary signals to guide diffusion toward hair structures consistent with both.

% The two conditions play deliberately different roles. The mesh is a dense but low-frequency descriptor: it pins down the approximate hair length, volume, and silhouette, yet says nothing about how fibers flow inside that volume. The sketch is the converse, a sparse but high-frequency descriptor of local direction and styling intent. Aligning two modalities this dissimilar, one volumetric and one projected, so that a single generated sample satisfies both is not trivial, and the design below is what makes it work.

\paragraph{Mesh conditioning}
To obtain a compact and structured conditioning signal for the diffusion model, we encode the input mesh $M$ with a pretrained 3D mesh VAE encoder~\citep{chen2025dora, zhang20233dshape2vecset} and further compress the resulting latent tokens with a learned attention pooling module:
\begin{equation}
\tilde{c}_{\mathrm{mesh}} = P_{\mathrm{mesh}}\bigl(E_{\mathrm{mesh}}(M)\bigr),
\end{equation}
where $E_{\mathrm{mesh}}(M)\in\mathbb{R}^{N_m\times d_m}$ is the mesh latent produced by the frozen encoder and $\tilde{c}_{\mathrm{mesh}}\in\mathbb{R}^{N'_m\times d}$ is the pooled mesh condition consumed by the transformer. Here $N_m$ and $N'_m$ are the numbers of tokens before and after pooling, $d_m$ is the mesh latent dimension, and $d$ is the transformer hidden dimension. 
% Pooling matters because the raw mesh latent is long enough to dominate the joint conditioning sequence and suppress the sketch stream.
Rather than treating all mesh tokens equally, the pooling module learns to weight and aggregate mesh geometric features according to their relevance to the hair generation process.

\paragraph{Sketch conditioning}
Input sketches may come from general-purpose image-editing models or artists, resulting in diverse stroke styles and densities. We therefore seek features that capture local hair directions while remaining robust to appearance variations, and adopt DINOv2~\citep{oquab2023dinov2} as our feature backbone. 
We observe that shallow features are overly sensitive to stroke density and width, whereas deep features emphasize shared semantic content, treating patches with different hair directions similarly. We therefore concatenate features from several intermediate layers to balance directional sensitivity and appearance robustness.
Given $V$ sketches ${S^v}$ and their corresponding orthographic cameras ${C^v}$, we extract these features independently for each view, incorporate camera information, and fuse the resulting tokens across views using self-attention.

% The input sketches may be generated by general-purpose image-editing models or drawn by artists. Unlike physical measurements, they provide illustrative cues about hair structure and may vary substantially in stroke style and density, even for the same hairstyle. We therefore seek features that capture local hair flow while remaining robust to such appearance variations.
% % Given $V$ input sketches $\{S^v\}$ and their corresponding orthographic cameras $\{C^v\}$, we extract multi-layer DINOv2 features~\citep{oquab2023dinov2} independently for each view, incorporate camera parameters, and fuse the resulting tokens across views using self-attention.

% We use a frozen DINOv2 ViT-S/14 encoder and investigate features at different depths using multiple sketches generated for the same hairstyle. Shallow features vary substantially across stroke styles and densities, indicating sensitivity to low-level appearance. In contrast, deep features are more invariant across sketches but become less discriminative of patches with different local hair orientations, reflecting their stronger emphasis on semantic content. We therefore combine intermediate features from layers 3, 6, and 9 to form a descriptor $f_i^v \in \mathbb{R}^{3n_d}$ for patch $i$ in view $v$, where $n_d$ is the feature dimension of each layer. Additional quantitative analysis is provided in the supplementary material.

To inject camera parameters, we compute a 6D Pl\"ucker ray $r_{i}^{v}\in\mathbb{R}^{6}$ for the center of each patch under camera $C^{v}$. Pl\"ucker parameterization encodes a ray directly rather than a camera pose, so it is independent of any global coordinate convention and varies smoothly with the viewpoint. 
We apply separate learned projections to the DINOv2 feature and the Pl\"ucker ray encoding, and concatenate the resulting embeddings to form a camera-aware patch token: \begin{equation} t_i^v = [\phi_f(f_i^v);\ \phi_r(r_i^v)], \end{equation} where $\phi_f$ and $\phi_r$ are learned projection layers. 
We then flatten the tokens from all views and process them jointly with a transformer to enable cross-view reasoning: 
\begin{equation}
c_{\mathrm{sketch}} = F_{\mathrm{sketch}}(\mathcal{T}_{\mathrm{sketch}}),
\end{equation}
where $\mathcal{T}_{\mathrm{sketch}}=\{t_{i}^{v}\}_{v=1,\,i=1}^{V,\,N_p}$ is the flattened token sequence over all views $N_p$ is the number of patches per view and $c_{\mathrm{sketch}}\in\mathbb{R}^{(V N_p)\times d}$ is the final sketch conditioning. 
Cross-attention over $c_{\mathrm{sketch}}$ naturally accommodates a variable number of input views.

\subsection{Strand Diffusion Transformer}
\label{sec:dit}
We use a conditional diffusion transformer to generate hairstyles in the orientation VAE latent space. Mesh and sketch information are provided through a joint conditioning sequence:
\begin{equation}
\mathbf{c} = [\tilde{c}_{\mathrm{mesh}};\ c_{\mathrm{sketch}}],
\end{equation}
where concatenation is along the token dimension. Given noisy latent tokens $\mathbf{z}_t \in \mathbb{R}^{L\times d_z}$, the transformer integrates these conditions through cross-attention, with the time variable $t$ incorporated via adaLN modulation. The network predicts a latent velocity field $\hat{\mathbf{v}} \in \mathbb{R}^{L\times d_z}$ and is trained using the flow-matching objective described in \autoref{sec:train}.

\subsection{Training and Inference}
\label{sec:train}

\paragraph{Training Data}
We collect approximately 55K strand-based hairstyles from four publicly available datasets: 40K from DiffLocks~\citep{rosu2025difflocks}, 10K from Perm~\citep{he2025perm}, 2K from UniHair~\citep{zheng2024towards}, and 3K from HairCS~\cite{lu2026haircsreconstructingstrandbasedhair}. All hairstyles are aligned to a common coordinate frame. Following HairLRM~\citep{shen2026hairlrm}, we extract a hair surface mesh and orientation fields from the ground-truth 3D strands. We also develop a pipeline to synthesize 2D sketches of strands from multiple viewpoints and in diverse styles, as detailed in next ection.

\begin{figure}[t]
    \centering
    \includegraphics[width=0.9\linewidth]{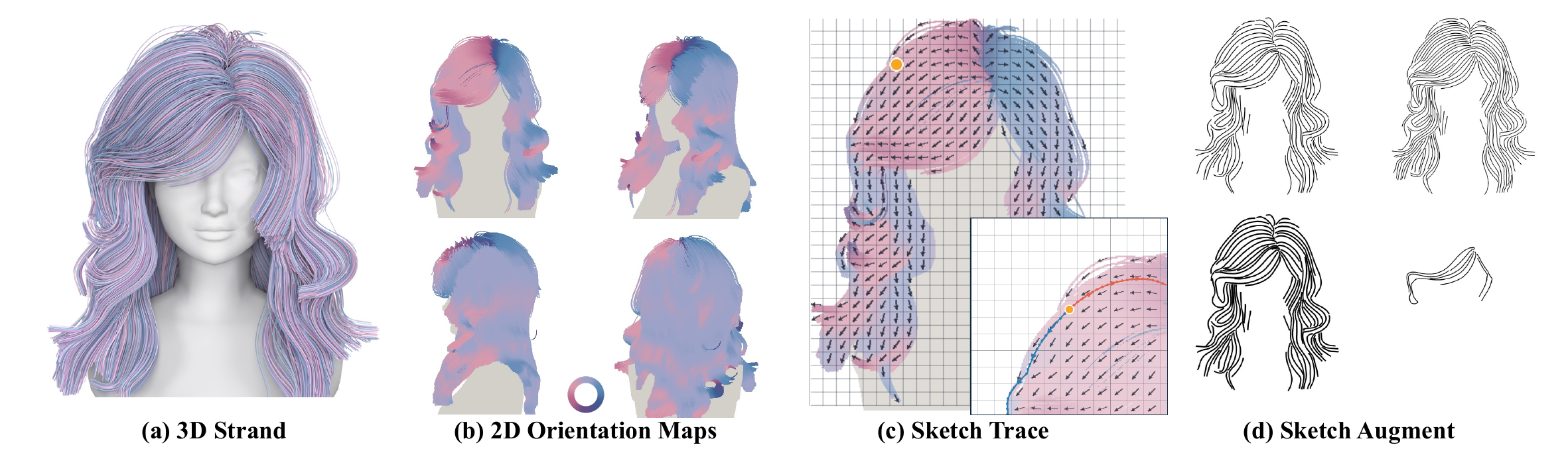}
    \vspace{-1em}
    \caption{
    Hair sketch synthesis pipeline. (a) Input strands and template head mesh. (b) Multi-view orientation maps; colors encode directed 2D tangents as shown in the circular legend. (c) Bidirectional tracing from sampled seeds; arrows show local directions, and the inset details tracing from the highlighted seed. (d) Sketch variations through stroke density, line width, and localized seed sampling.
}
  \vspace{-1em}
    \label{fig:sketchsynth}
\end{figure}
\paragraph{Sketch Synthesis}
At inference time, sketches can be generated by image-editing models or drawn by artists, but collecting them at scale with accurate 3D strand alignment is challenging. Converting strand renderings into sketches with image-editing models is also too time-consuming at scale. We therefore develop an efficient pipeline to synthesize aligned, diverse training sketches directly from 3D strands(\autoref{fig:sketchsynth}).

We first convert 3D strands into a 3D orientation field represented by points cloud. For each viewpoint, we project these directions onto the image plane and splat the points to obtain a 2D orientation map, using the head depth map to resolve hair–head occlusion. 
We trace strokes bidirectionally from uniform grid cell centers along local hair orientations, stopping at the background, the head region, or sharp directional changes that typically indicate hair partings. 
To promote even coverage, we maintain an occupancy map on the same grid, reject strokes that cross occupied cells, and mark cells traversed by accepted strokes as occupied. 
We also remove very short strokes to suppress noise. 

To augment sketch diversity, we vary grid size to control stroke density, adjust stroke width and smoothness, and trace from a small subset of grid cells to generate partial sketches. We also randomly sample viewpoints with a bias toward frontal views. Together, these augmentations generate diverse sketches from limited ground-truth hairstyles while preserving their underlying geometry.

\paragraph{Orientation VAE Training}
We train the orientation VAE with orientation reconstruction loss and KL divergence term,
\begin{equation}
L_{\mathrm{VAE}} = L_{\mathrm{ori}} + \lambda_{\mathrm{kl}} L_{\mathrm{kl}},
\end{equation}
where $L_{\mathrm{ori}} = \frac{1}{N}\sum_{i=1}^{N}\|\hat{\mathbf{o}}_i-\mathbf{o}_i\|_2^2$ is the distance between predicted and ground-truth orientations at sample points inside the hair region. We set $\lambda_{\mathrm{kl}}$ to a small value to promote reconstruction accuracy. 

Following HairLRM, we prioritize sampling rare hair orientations. We additionally use local orientation variance to sample regions with rapid directional changes more frequently.
As training progresses, we gradually increase the sampling probabilities of both rare orientations and high-variance regions to improve reconstruction of complex hairstyles.

\paragraph{Diffusion Training}
We train the diffusion transformer in the orientation VAE latent space using flow matching. Given a latent code $\mathbf{z}_{\mathrm{data}}$, Gaussian noise $\boldsymbol{\epsilon}\sim\mathcal{N}(\mathbf{0},\mathbf{I})$, and time $t\sim\mathcal{U}(0,1)$, we define the interpolated latent as $\mathbf{z}_t=(1-t)\boldsymbol{\epsilon}+t\mathbf{z}_{\mathrm{data}}$. The network predicts the target velocity $\mathbf{z}_{\mathrm{data}}-\boldsymbol{\epsilon}$, conditioned on $\mathbf{c}$:
\begin{equation}
L_{\mathrm{FM}}
=
\mathbb{E}_{t,\mathbf{z}_{\mathrm{data}},\boldsymbol{\epsilon},\mathbf{c}}
\left[
\left\|
\mathbf{v}_{\theta}(\mathbf{z}_t,t,\mathbf{c})
-
(\mathbf{z}_{\mathrm{data}}-\boldsymbol{\epsilon})
\right\|_2^2
\right].
\end{equation}
We independently drop mesh and sketch conditions during training, using a null token for the dropped mesh condition. At inference, we apply decoupled classifier-free guidance:
\begin{equation}
\label{eq:cfg}
\mathbf{v}_{\mathrm{cfg}}
=
\mathbf{v}_\varnothing
+
s_m(\mathbf{v}_m-\mathbf{v}_\varnothing)
+
s_s(\mathbf{v}_s-\mathbf{v}_\varnothing),
\end{equation}
where $\mathbf{v}_\varnothing$, $\mathbf{v}_m$, and $\mathbf{v}_s$ are the unconditional, mesh-only, and sketch-only predictions, respectively, and $s_m$ and $s_s$ control the mesh and sketch guidance strengths. Starting from Gaussian noise at $t=0$, we integrate the guided velocity to $t=1$ to generate a latent code.

To improve robustness, we add Gaussian noise ($\sigma=0.1$) to the input mesh latents to reduce overfitting to the limited training data. We also perturb sketch camera directions with Gaussian jitter ($\sigma=5^\circ$) to simulate camera estimation errors at inference time. 
At each training iteration, we sample a variable number of sketch views to support both single-view and multi-view inputs, and randomly select a precomputed augmented sketch for each view.

\paragraph{Inference}
For an in-the-wild image, we use an image generation model~\citep{Google2026NanoBanana} to synthesize corresponding multi-view images. We remove texture and illumination, decompose the resulting images into hair-only renderings, and reconstruct the hair meshes using an image-to-3D model~\citep{Tripo3D2026H31}. We also extract hair-flow sketches from the input image. 
Alternatively, artists may draw directional strokes directly on a coarse 3D hair mesh as guidance; we project these strokes into the specified camera views to obtain posed sketches.
We transform the mesh and sketches into spatially aligned conditions following the coordinate and camera conventions used during training. Given these conditions, we sample a strand latent, decode it into a volumetric orientation field, and recover explicit strands following a tracing procedure similar to CT2Hair~\citep{shen2023ct2hair}. Further details are provided in the supplementary material.
% For in-the-wild inference, we use image generation model~\citep{Google2026NanoBanana} to convert a monocular image into multi-view images, then remove color and lighting to obtain clay renderings. 
% We isolate the hair while retaining the full-head renderings and use them as input to a 3D generation model~\citep{Tripo3D2026H31} to obtain the full head mesh and the hair mesh. We also extract a sketch from the original image. Prompts and details are provided in the supplementary material.
% Then we register the hair mesh to the full-head mesh using ICP, then align both to the template bust in the orientation VAE’s training coordinate system. This alignment uses 3D landmarks obtained by back-projecting 2D detections from multi-view mesh renderings. We then fit an orthographic camera using 2D facial landmarks to align the full-head mesh with the original image and project the hair mesh with the same camera. 
% By prompting sketch extraction to preserve the image’s spatial layout, the projected hair mesh and sketch are approximately aligned.
% Given the aligned mesh and posed sketches, we sample a strand latent with the trained diffusion transformer using the decoupled guidance of \autoref{eq:cfg}, decode it into a dense volumetric orientation field by evaluating \autoref{eq:decoder} on query points inside the hair mesh, and extract explicit strands from that field with CT2Hair~\citep{shen2023ct2hair}.

\section{Experiments}
\label{sec:experiments}
\begin{figure*}[t]
    \centering
    \includegraphics[width=0.9\linewidth]{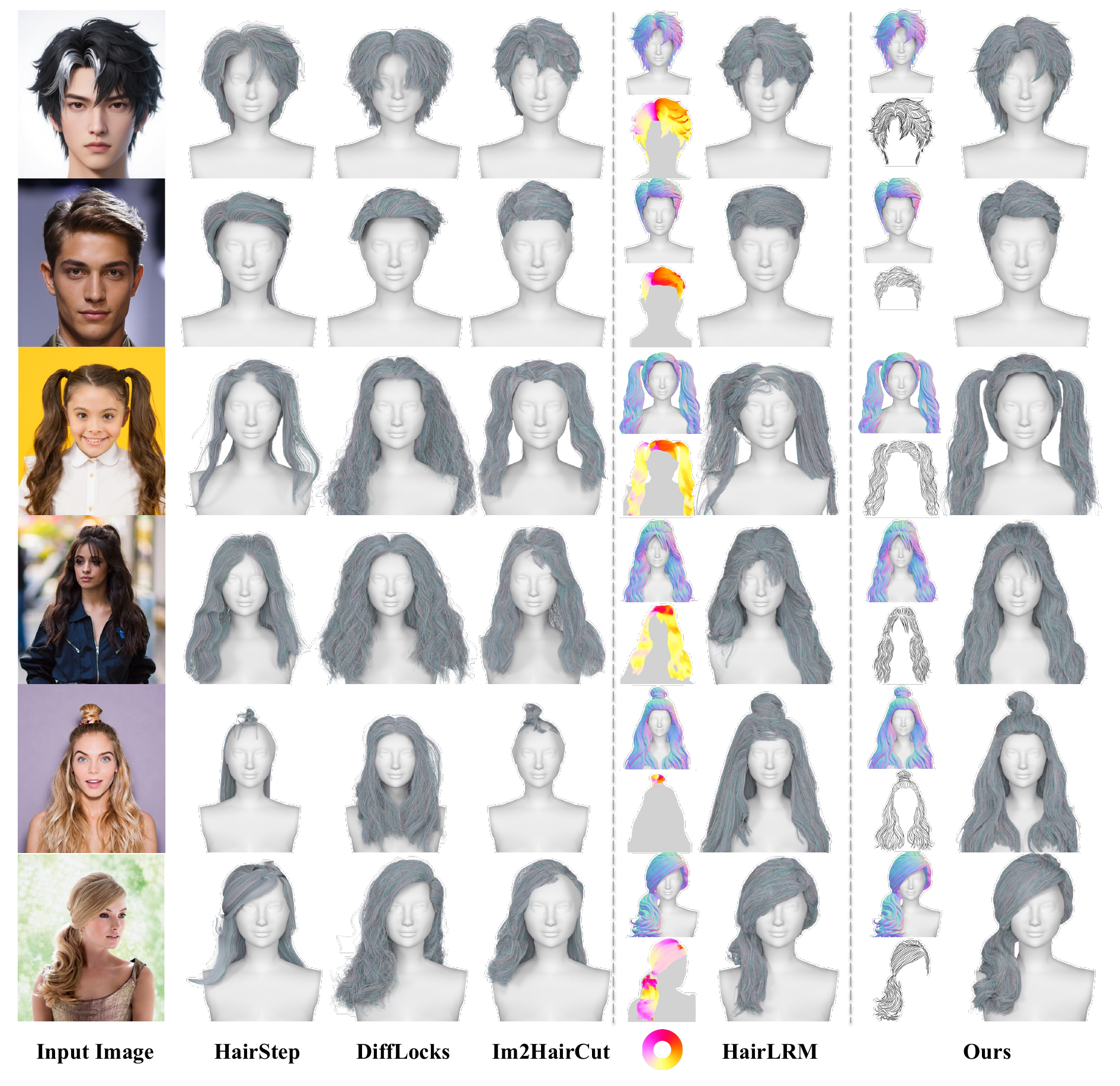}
    \vspace{-1.5em}
    \caption{Qualitative comparison of single-view 3D hair reconstruction. HairLRM and HairFlow share the same input mesh, supplemented by HairStep orientation maps or sketches from image generation models, respectively. Inputs appear to the left of the corresponding strands. The color wheel below HairLRM illustrates the direction-to-color mapping.}
  \vspace{-1.5em}
    \label{fig:comparison}
    
\end{figure*}

% We evaluate \ours along three axes that follow directly from our claims: the capacity of the orientation-only strand latent space (\autoref{tab:vae_recon}), the fidelity of generated hairstyles against state-of-the-art strand modeling methods (\autoref{tab:synthetic}, \autoref{fig:hairgpt}), and the degree to which the mesh and sketch conditions provide separable control (\autoref{tab:control}, \autoref{fig:sketch_ablate}). We additionally report runtime (\autoref{tab:runtime}), examine robustness to sketch style and to the number of sketch views, and demonstrate a downstream simulation application.

\paragraph{Implementation details}
Our StrandVAE contains 48M parameters and represents each hairstyle as a $2048\times64$ latent tensor. We train it for 200 epochs with a global batch size of 192 using AdamW with a learning rate of $5\times10^{-4}$. Our conditional DiT contains 16 transformer blocks with a hidden dimension of 512 and 8 attention heads. We encode the input hair mesh using the frozen encoder of a pretrained DORA-VAE~\citep{chen2025dora}. Its 2,048 mesh tokens are reduced to 1,024 tokens through learned attention pooling. For each $336\times336$ sketch, we extract 576 patch tokens using a frozen DINOv2 ViT-S/14~\citep{oquab2023dinov2} and concatenate features from layers 3, 6, and 9. We train the DiT for 300 epochs on four GPUs with a per-GPU batch size of 80 and a learning rate of $3.5\times10^{-4}$. During training, we randomly use 1, 2, or 4 sketch views. At inference, we use 30 Euler steps and extract explicit strands from the generated orientation field following a tracing procedure similar to CT2Hair~\citep{shen2023ct2hair}. Further details are provided in the supplementary material.

% \paragraph{Evaluation protocol and metrics}
% We report two families of metrics. \emph{Orientation error} (Ori-Err, degrees, lower is better) is the mean angle between the predicted and ground-truth 3D hair orientation at query points sampled inside the hair region, and is the metric that directly measures what our latent space represents. \emph{Silhouette IoU} (higher is better) is computed between the hair mask of the rendered result and that of the ground truth under the input view, and measures whether the global shape is correct. Because our sampler is stochastic, every number we report for \ours is averaged over several samples per input drawn with different seeds, and the mesh and sketch conditions are held fixed within an input. All 3D results, ours and baselines alike, are registered to a common bust before measurement. 
% \todo{confirm the two metric definitions, state the number of samples per input and the seed handling, and decide whether to add a strand-level metric (e.g.\ Chamfer distance between strand point sets) so that the evaluation is not purely field-based.}

\begin{table*}
    \centering
    \caption{Quantitative comparison. Silhouette IoU ($\uparrow$) is evaluated on 30 real images. Angular error ($^\circ$, $\downarrow$) is measured between 3D orientation fields derived from reconstructed and ground-truth strands on 326 validation hairstyles.
    }
    \label{tab:compare}
    \begin{tabular}{lcccc>{\columncolor{green!8}}c}
        \toprule
        Metric & HairStep & DiffLocks & Im2Haircut & HairLRM & \textbf{HairFlow (Ours)} \\
        \midrule
        IoU $\uparrow$    &  0.625 & 0.583& 0.773& 0.855&  0.912\\
        Ori-Err $\downarrow$ & $56^\circ$& $43^\circ$& $35^\circ$& $29^\circ$& $22^\circ$\\
        \bottomrule
    \end{tabular}
    \vspace{-1em}
\end{table*}

%%%%%%%%%%%%%%%%%%%%%%%%%%%%%%%%%%%%%%%%%%%%%%%%%%
\subsection{Comparisons}
%%%%%%%%%%%%%%%%%%%%%%%%%%%%%%%%%%%%%%%%%%%%%%%%%%
We compare HairFlow with HairStep~\cite{zheng2023hairstep}, DiffLocks~\cite{rosu2025difflocks}, Im2Haircut~\cite{sklyarova2025im2haircut}, and HairLRM~\cite{shen2026hairlrm}. 
HairStep lifts predicted strand and depth maps into 3D, while DiffLocks generates strand using image-conditioned diffusion over scalp-uv parameterization. 
Im2Haircut combines a transformer-based hair prior with Gaussian-splatting-based optimization. 
HairLRM optimizes the latent code of a pretrained hair autoencoder using a generated mesh and a HairStep strand map. Our method uses the same mesh and a sketch extracted by large image generation model as explicit geometric and directional constraints. Comparisons with HairGPT~\cite{luo2026hairgpt} and \cite{hairlrm2_conghui} are in the supplementary material.

\paragraph{Qualitative comparison}
Qualitative results can be found in in Fig~\ref{fig:comparison}.
HairStep, Im2Haircut, and DiffLocks are trained on limited hair datasets and thus generalize poorly to complex real-world hairstyles, such as half-up buns and side ponytails. HairLRM involves meshes generated by a 3D foundation model, but still relies on 2D cues predicted by HairStep, which can be inaccurate for complex hairstyles, as shown in the final two rows. Moreover, HairLRM fits a compact  VAE latent to match the input mesh, which may not fully reproduce its geometry, leading to strand silhouettes visibly deviating from the input. In contrast, our method generalizes to diverse hairstyles while closely adhering to the geometry and strand flow specified by the input mesh and sketch.

\paragraph{Quantitative comparison}
We collect 30 real images and compute IoU between rendered strand silhouettes and input hair segmentation. For 3D evaluation, we hold out 326 validation hairstyles from training datasets, render them with a template head in Blender at angles from $-30^\circ$ to $30^\circ$, and use Nano Banana to complete the faces while keep the hair unchanged. The resulting images serve as inputs to baselines. At each spatial sample, we obtain directions from the nearest reconstructed strand point and the nearest ground-truth strand point, then average angular error. The quantitative results can be found in Table~\ref{tab:compare}

\begin{figure}[t!]
    \centering
    \includegraphics[width=0.85\linewidth]{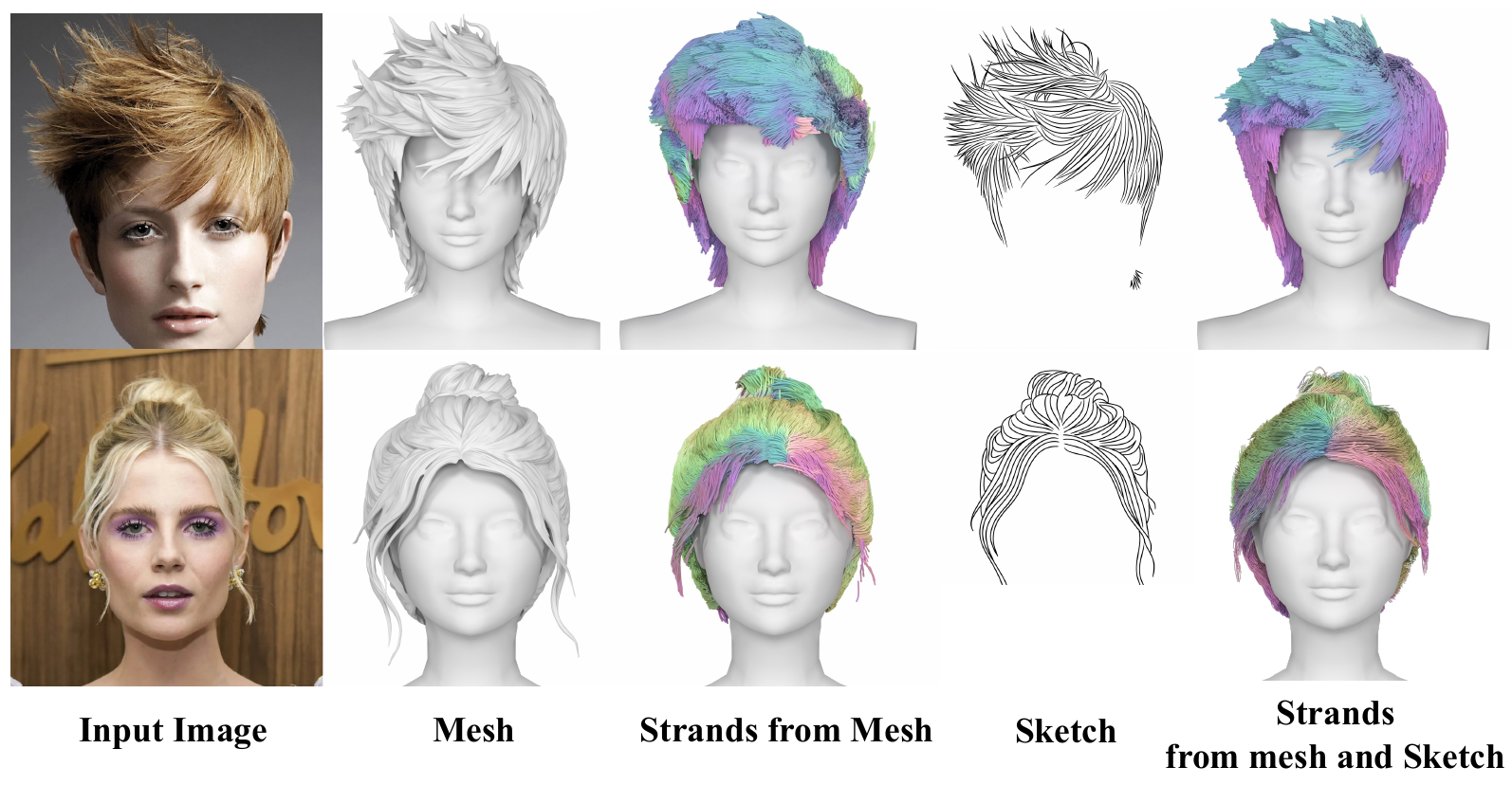}
    \vspace{-1em}
    \caption{
    \textbf{Effect of the sketch condition.}
     From left: input image, mesh, mesh-only result, sketch, and mesh+sketch result. Results within each row share the same mesh and random seed. Colors encode 3D tangent directions.
    }
    \vspace{-1em}
    \label{fig:sketch_ablate}
\end{figure}

\begin{table}[t]
    \centering
    \caption{
    Additional quantitative comparisons with HairLRM. We report mean angular errors ($^\circ$, lower is better) for VAE reconstruction on the training and test sets, and for strand generation conditioned on meshes alone or with additional 2D cues.
    }
    \label{tab:vae_recon}
    \begin{tabular}{lcccc}
        \toprule
         Method & Training set & Validation set & Mesh Only & Mesh + 2D cues \\
         \midrule
         HairLRM~\citep{shen2026hairlrm} & 10.47 & $12.70$ & $36.81^\circ$& $29.52^\circ$\\
         \rowcolor{green!8}
         \textbf{HairFlow (Ours)} & $\mathbf{4.01}$ & $\mathbf{6.25}$ &$27.43^\circ$ & $22.92^\circ$\\
         \bottomrule
    \end{tabular}
    \vspace{-1em}
\end{table}

\paragraph{Inference Time}
We benchmark available implementations and quote reported runtimes for HairGPT and \citet{hairlrm2_conghui}, whose public code we could not obtain. HairStep, DiffLocks, and Im2Haircut take 30.7, 48.8, and 378.3 s per image, respectively; HairGPT reports 30--60 s. HairLRM, \citet{hairlrm2_conghui}, and HairFlow additionally use foundation models to obtain geometric conditions. HairLRM takes $\sim$15 min: 3 min for mesh generation, 10 min for latent optimization, and 1--2 min for strand growing. \citet{hairlrm2_conghui} reports $\sim$20 min, including multi-view image editing and $\sim$6 min for semantic processing and orientation propagation. HairFlow requires $\sim$3 min for mesh and sketch generation, 6.9 s for condition encoding, diffusion sampling, and field decoding, and 1--2 min for strand growing. Its core generation stage is approximately $87\times$ faster than HairLRM's reported latent optimization.

\paragraph{More quantitative comparison with HairLRM}
\autoref{tab:vae_recon} provides additional quantitative comparisons with HairLRM~\citep{shen2026hairlrm}. Our Orientation VAE reduces reconstruction angular error from $10.47^\circ$ to $4.01^\circ$ on the training set and from $12.70^\circ$ to $6.25^\circ$ on the validation set.
This improvement benefits from a larger latent representation devoted entirely to the orientation field, without sharing capacity with SDF modeling.
For mesh-conditioned generation on validation set, HairFlow achieves lower angular error both with and without 2D cues. Adding sketches further reduces our error from $27.43^\circ$ to $22.92^\circ$, compared with HairLRM's $29.52^\circ$.

\subsection{Ablations}
%%%%%%%%%%%%%%%%%%%%%%%%%%%%%%%%%%%%%%%%%%%%%%%%%%

\paragraph{Conditioning Ablation}
We compare mesh-only conditioning with joint mesh and sketch conditioning in \autoref{fig:sketch_ablate}. The mesh constrains the overall hair shape but does not explicitly specify strand directions. Adding sketches improves alignment with the hair flow in the input images, particularly in the second rows. Sketches also provide finer structural cues, producing more clearly separated hair clumps in the first row.
The quantitative results in Table~2 further confirm that adding 2D sketches reduces angular error.

\begin{figure}[t!]
    \centering
    \includegraphics[width=0.9\linewidth]{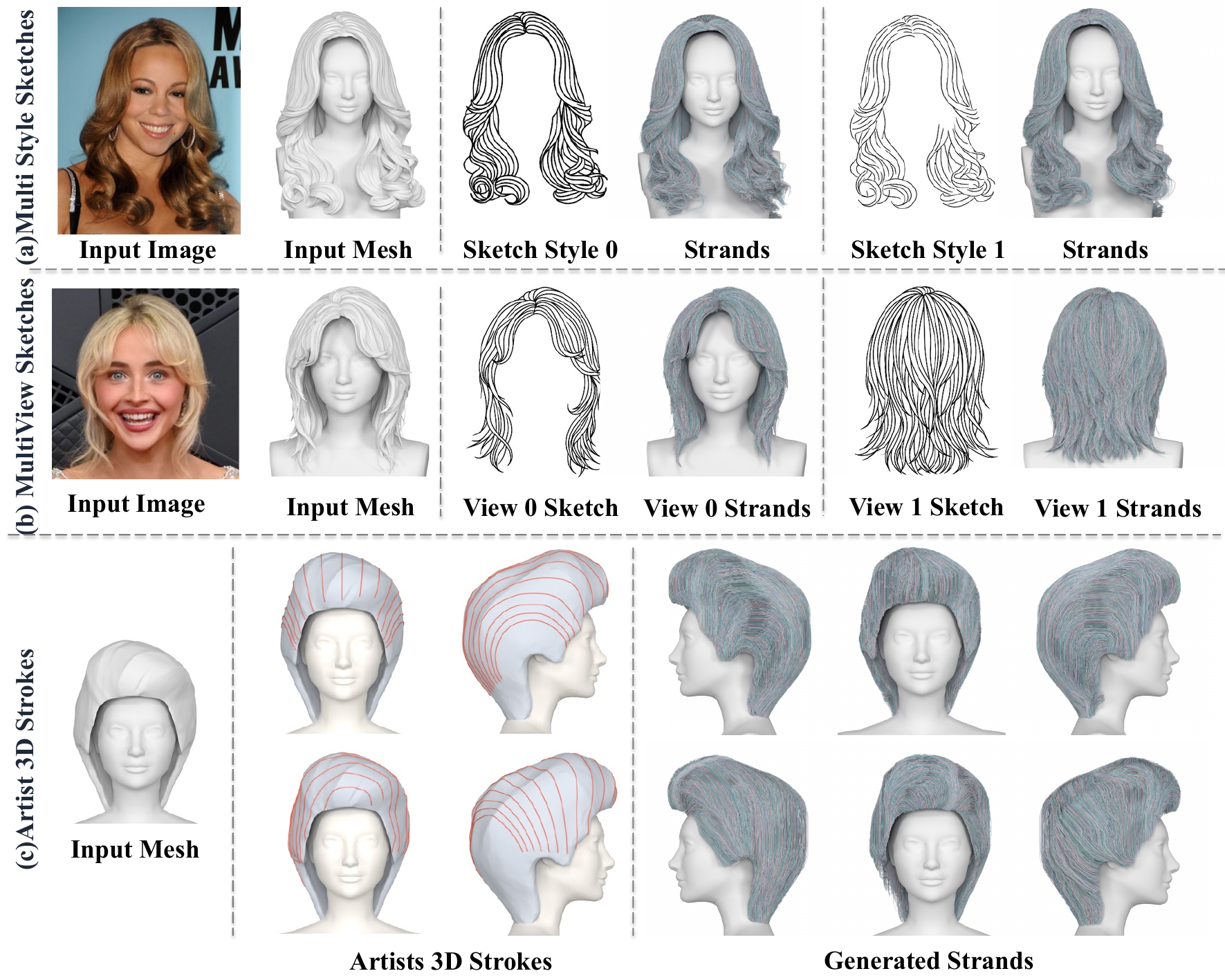}
    \vspace{-0.5em}
    \caption{
    \textbf{Flexible Sketch Guidance:}
    (a) Sketches with different stroke widths and styles yield similar strand geometry. (b) Joint conditioning on sketches from two viewpoints produces a single hairstyle consistent with both views. (c) Two sets of artist-drawn 3D strokes on the same mesh are projected into multi-view sketches, guiding the generated strands to follow different flow patterns.
    }
    \vspace{-1em}
    \label{fig:gallery}
\end{figure}

\paragraph{Flexible Sketch Guidance}
\ref{fig:gallery} illustrates the versatility of sketch guidance in HairFlow.
In (a), two sketches of the same hairstyle differ substantially in appearance, yet produce similar strand structures, indicating robustness to stylistic variations while preserving local flow. In (b), front and back sketches jointly condition a single generation, yielding strands that follow the directional cues in both views. In (c), we project artist-drawn 3D strokes into multi-view sketches and use them as conditions. With the same input mesh, two sets of strokes produce distinct strand flow patterns aligned with the artist’s intent, demonstrating directional control within a fixed global shape.

\section{Conclusion}
\label{sec:conclusion}
We presented HairFlow, a multimodal conditional diffusion framework for 3D hair strand modeling. It combines mesh--sketch cues from foundation models or artists with learned 3D strand priors to generate orientation fields without per-instance latent optimization. Meshes constrain global shape, while sketches guide local flow. An automatic sketch synthesis pipeline provides geometrically aligned training pairs. Experiments demonstrate improved input alignment, reduced inference time, and flexible control across modalities and styles.

\bibliographystyle{iclr2027_conference}
\bibliography{main}

\appendix
\section{More Experiments}
\paragraph{More Comparisons}
We provide additional comparisons with HairGPT~\citep{luo2026hairgpt} and \citet{hairlrm2_conghui}. HairGPT autoregressively generates strand sequences, organizing hairstyles by semantic scalp regions and hierarchical strand structure. \citet{hairlrm2_conghui} extract directional and semantic cues from multi-view renderings of an LRM-generated mesh, then propagate surface orientations into the hair volume using hand-crafted rules.
Our generated strands follow the input hairstyles more faithfully than those of HairGPT and achieve visual quality comparable to \citet{hairlrm2_conghui}. Both our method and \citet{hairlrm2_conghui} incur additional costs for 3D generation and image editing to obtain geometric conditions. Their pipeline reportedly takes approximately 20 min per example. In our pipeline, condition generation and strand growing together take approximately 4–5 min, while the core diffusion sampling and orientation-field decoding require only 6.9 s.

\paragraph{Hair simulation}
To demonstrate the downstream usability of our generated strands, we import the reconstructed hairstyles into Unreal Engine and simulate their motion under wind. This experiment illustrates how our static reconstructions can be incorporated into a strand-based animation pipeline, extending their use beyond static rendering. The supplementary video shows the resulting hair motion.

\begin{figure}
    \centering
    \includegraphics[width=\linewidth]{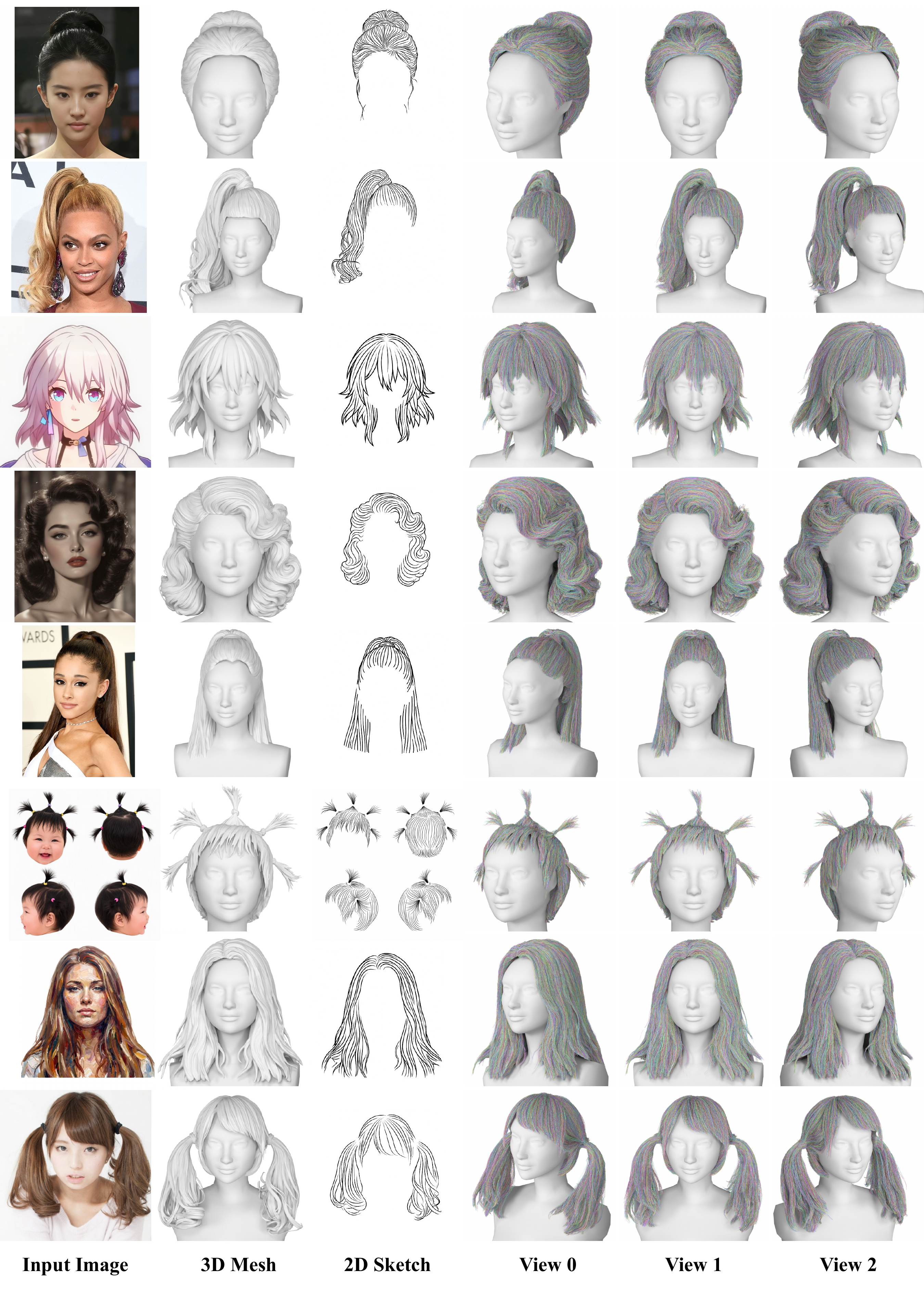}
    \caption{Multi-view renderings of reconstructed strands. Rows 3 and 7 use anime and oil-painting inputs, respectively; row 6 uses multi-view image inputs. The results demonstrate reconstruction across diverse hairstyles, image styles, and input views.}
    \label{fig:cases_0}
\end{figure}

\begin{figure}
    \centering
    \includegraphics[width=\linewidth]{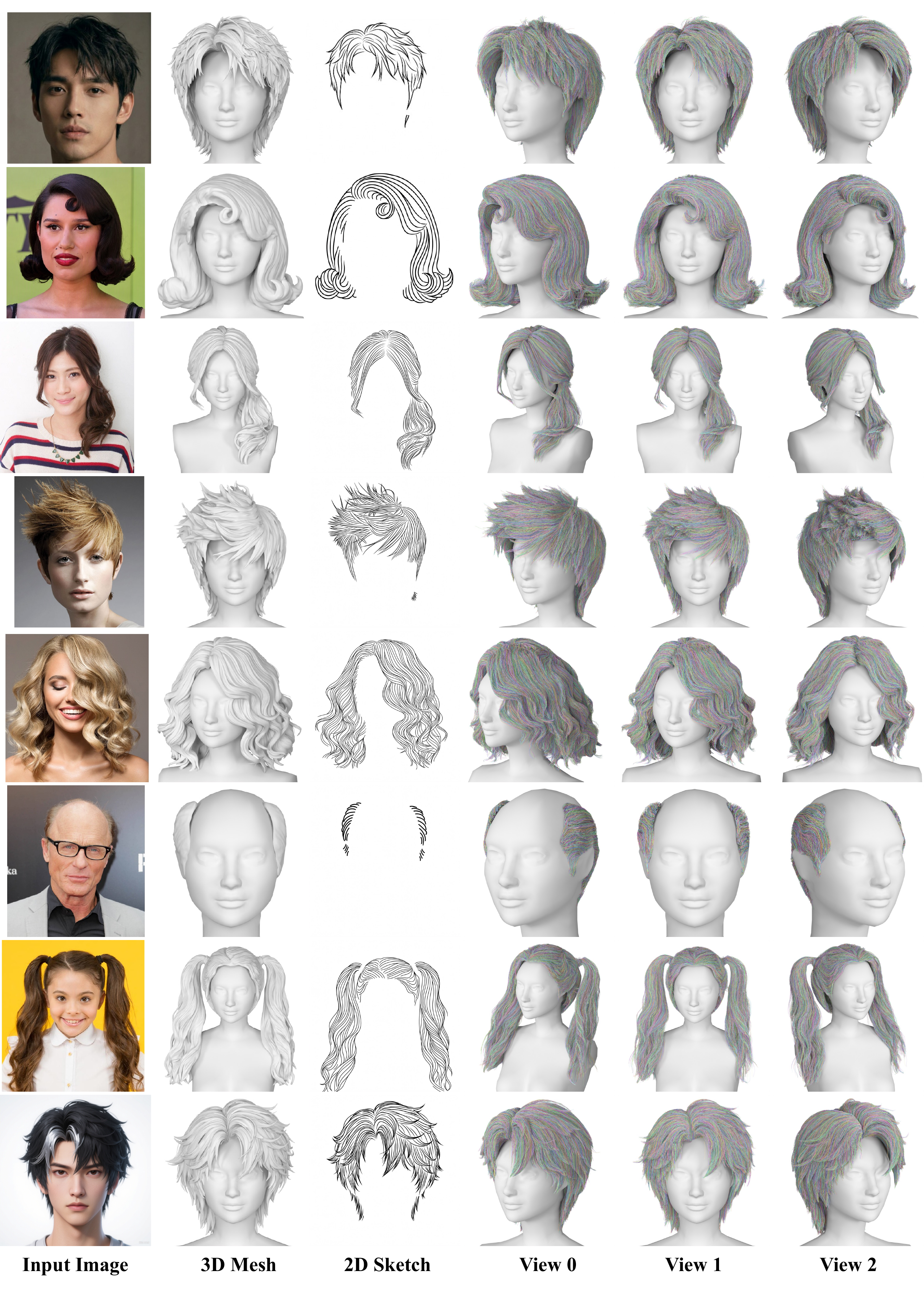}
    \caption{Additional multi-view renderings of reconstructed strands, showing hairstyle geometry and strand flow from different viewpoints.
}
    \label{fig:cases_1}
\end{figure}

\begin{figure}
    \centering
    \includegraphics[width=\linewidth]{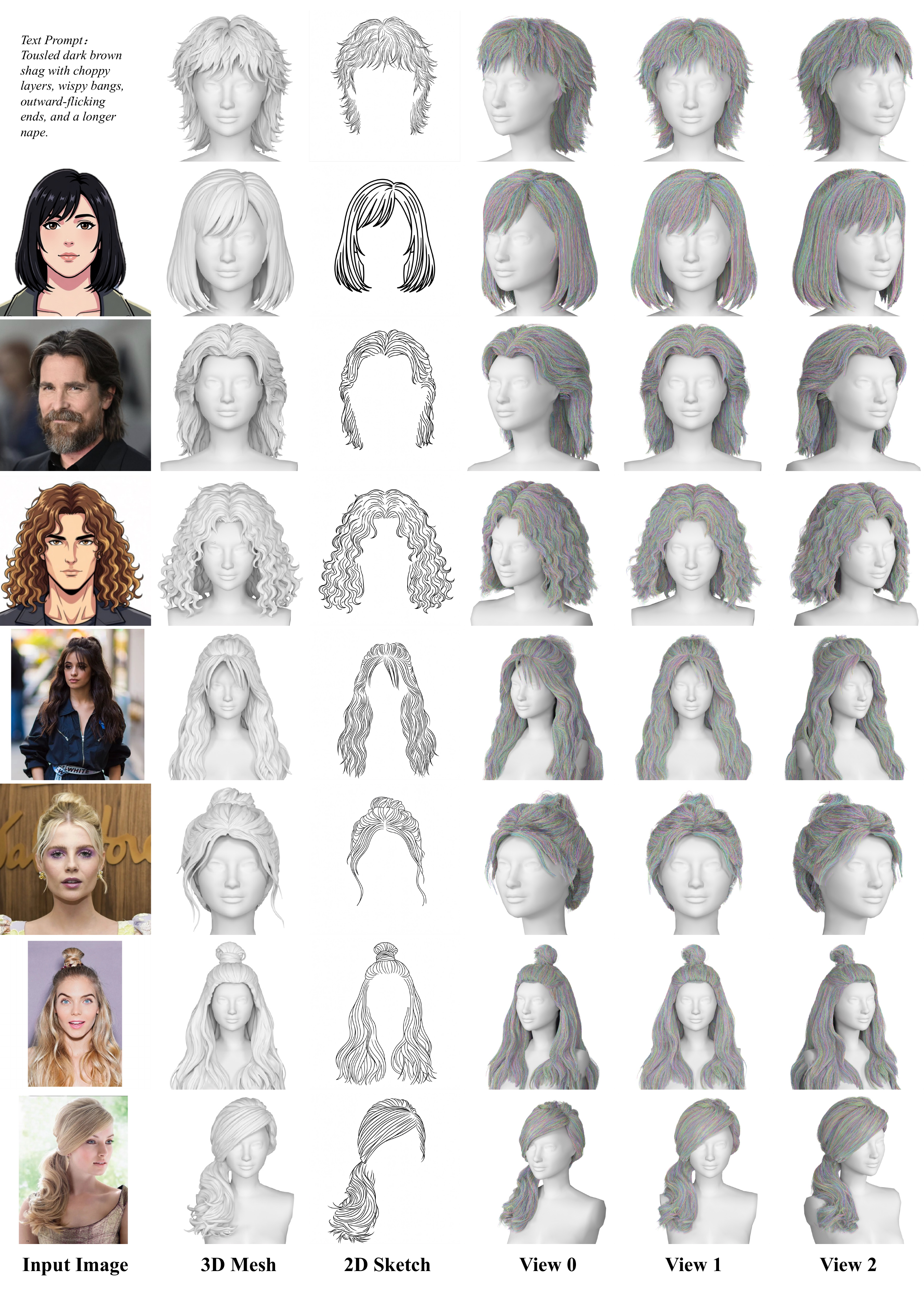}
    \caption{Additional multi-view results. In the first row, an artist-provided text prompt is used to generate a hair mesh, and sketches are extracted from its rendered images, forming our unified geometric interface for strand generation.
}
    \label{fig:cases_2}
\end{figure}

\paragraph{Multi-view visualization}
We provide additional renderings of reconstructed strands from multiple viewpoints in \autoref{fig:cases_0}, \autoref{fig:cases_1}, and \autoref{fig:cases_2}. These views reveal the overall hairstyle geometry and strand flow beyond the input viewpoint, allowing closer inspection of the reconstructed 3D structure.

  \section{Additional Implementation Details}
  \label{sec:supp_impl}

  \begin{figure}
    \centering
    \includegraphics[width=\linewidth]{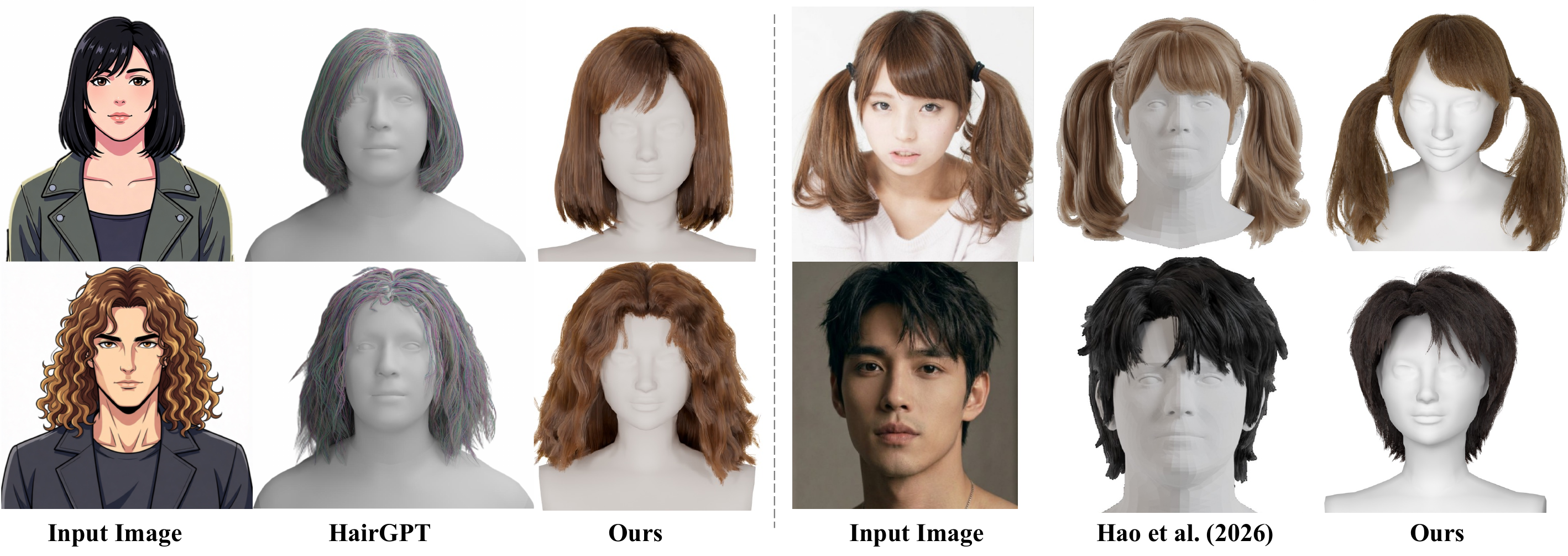}
    \caption{Qualitative comparisons with HairGPT~\citep{luo2026hairgpt} and \citet{hairlrm2_conghui}. Their input images and reconstruction results are taken from the respective papers, as code is unavailable.}
    \label{fig:more_compare}
\end{figure}

  \paragraph{Orientation VAE.}
  The orientation vae uses a hidden dimension of 384. Its encoder and latent transformer contain
  8 and 16 layers, respectively, with 6 attention heads. For each training sample, we draw
  32,768 orientation points and 32,768 salient points as encoder inputs, and supervise the
  decoder at 3,000 query points. The query points are sampled according to a combination of
  inverse orientation frequency and local directional variance. We use an orientation
  reconstruction loss together with a latent regularization term weighted by $10^{-4}$.
  Training uses cosine learning-rate decay and gradient clipping with a maximum norm of
  1.0.

  \paragraph{Condition encoders.}
  For mesh conditioning, we use the frozen encoder of the pretrained DORA-VAE-1.1 model,
  which produces 2,048 tokens of dimension 64. A learned attention-pooling module reduces
  them to 1,024 tokens. For sketch conditioning, we resize each view to $336\times336$ and
  use a frozen DINOv2 ViT-S/14. We concatenate its features from layers 3, 6, and 9. Each
  view contains $24\times24=576$ patch tokens. A 6D Pl\"ucker ray is computed for every
  patch and projected to a 32-dimensional camera embedding. Tokens from all views are
  jointly processed by a 6-layer transformer.

  \paragraph{Training augmentation.}
  The mesh latent is perturbed with Gaussian noise of standard deviation 0.1. We
  independently perturb the camera azimuth and elevation with Gaussian noise of standard
  deviation $5^\circ$. Mesh and sketch conditions are independently dropped with
  probability 0.1. At each iteration, the number of input sketch views is sampled from $
  \{1,2,4\}$.

  \paragraph{Sketch synthesis.}
  We render eight $512\times512$ orthographic orientation maps for each training hairstyle
  and trace them bidirectionally to produce sketches. We vary stroke spacing and width by
  sampling the occupancy-cell size from $\{8,12,16\}$ pixels and the stroke width from $
  \{1,2,3\}$ pixels. We also generate partial sketches by restricting stroke seeds to
  randomly sampled local regions.

  \paragraph{Sampling and strand extraction.}
  We scale Orientation VAE latents by 0.57 during diffusion training and undo this scaling after
  sampling. We use 30 Euler steps and decoupled classifier-free guidance:
  \begin{equation}
      \mathbf{v}_{\mathrm{cfg}}
      =
      \mathbf{v}_{\varnothing}
      +
      s_m(\mathbf{v}_m-\mathbf{v}_{\varnothing})
      +
      s_s(\mathbf{v}_{m,s}-\mathbf{v}_m),
  \end{equation}
  where $\mathbf{v}_{\varnothing}$, $\mathbf{v}_m$, and $\mathbf{v}_{m,s}$ are the
  unconditional, mesh-conditioned, and jointly conditioned predictions, respectively. We
  use $s_m=1.5$ and $s_s=5.0$. The sampled latent is decoded on a regular grid inside the
  hair mesh. Explicit strands are then recovered from the volumetric orientation field
  using a tracing procedure similar to CT2Hair~\citep{shen2023ct2hair}.

\section{Input Construction and Alignment}
\label{sec:supp_input}

\paragraph{Input construction from a single image.}
Given an in-the-wild image, we first use an image generation model~\citep{Google2026NanoBanana} to synthesize front, back, left, and right views while preserving the identity and hairstyle of the reference. We then remove texture and illumination from the generated views to obtain clay-style renderings. These renderings are further decomposed into full-head and hair-only images. We feed the two sets of renderings to an image-to-3D model~\citep{Tripo3D2026H31} to reconstruct a full-head mesh and a coarse hair mesh, respectively. Hair-flow sketches are also extracted from the reference image or generated views while preserving their image-space layouts.

\paragraph{Mesh registration.}
Because the head and hair meshes are reconstructed independently, we first register the hair mesh to the full-head mesh using similarity ICP. We then align both meshes to the template bust in the coordinate system of the Orientation VAE. Specifically, we render the reconstructed head from the frontal and two nearby views, detect facial landmarks in the rendered images, and back-project them onto the mesh. A similarity transformation is estimated between these 3D landmarks and the corresponding landmarks on the template bust and is applied to both the head and hair meshes.

\paragraph{Camera and sketch alignment.}
We fit an orthographic camera by matching facial landmarks between the reconstructed head and the input view. For multi-view inputs, the remaining cameras are obtained by rotating the fitted frontal camera around the vertical axis. Image generation may introduce small view-dependent changes in scale and framing. We therefore estimate an additional 2D similarity transformation for each view by matching the projected hair-mesh silhouette with the corresponding hair-only rendering. The same transformation is applied to the hair-flow sketch. Finally, the aligned sketches are resized and arranged according to the camera convention used during training.

\paragraph{Artist-provided guidance.}
An artist can draw directional strokes directly on a coarse 3D hair mesh. When the mesh is transformed into the canonical Orientation VAE coordinate system, the same transformation is applied to the strokes. We then project the strokes through user-specified or automatically placed orthographic cameras and rasterize them as hair-flow sketches. Since the strokes are attached to the mesh surface and projected with the same cameras, the resulting mesh and sketch conditions are aligned by construction. The artist may draw strokes only in selected regions or provide guidance from a subset of views.

\paragraph{Image-generation prompts.}
We use task-specific prompts for multi-view generation, appearance removal, component decomposition, and sketch extraction. The multi-view prompt asks the image model to preserve the subject and hairstyle of the reference image, infer occluded hair regions, and follow a predefined four-view layout. A version of the prompt is:

\begin{quote}\small\itshape
\textbf{Multi-View Generation}: Reference Image 1 is the sole source of the character’s identity, hairstyle, and visual style. Reference Image 2 is a four-panel template used only to control the viewpoints, head orientations, positioning, and layout.
Generate four head views of the person from Reference Image 1: front, back, left profile, and right profile. Accurately infer and reconstruct the hair regions occluded in Reference Image 1. Preserve the facial identity, hairstyle, and visual style of Reference Image 1 in the final output.
Use a pure white background. Remove everything below the neck and any objects or accessories on the head. Match the viewpoints, positioning, and four-panel layout of Reference Image 2 exactly. Maintain strict identity, hairstyle, and geometric consistency across all four views.
\end{quote}

\begin{quote}\small\itshape
\textbf{Clay Conversion}: Generate a square four-panel grid showing a gray 3D clay sculpt. The sculpted details should be fine, smooth, and rich, with no rough surface texture. Use a pure white background. Match the geometry of the reference image exactly, while removing its original colors and photorealistic lighting and shadows. Keep the output dimensions identical to the reference image, and strictly align the position and viewpoint of each panel with the reference.
\end{quote}

\begin{quote}\small\itshape
\textbf{Hair Isolation}: Isolate the hair in each panel of the four-panel reference image and remove all other elements. Keep the hair in exactly the same position, with the original viewpoints and four-panel layout unchanged.
\end{quote}

\begin{quote}\small\itshape
\textbf{Sketch Extraction}: Input: A portrait image.
Output: A hair geometry path map with the same dimensions as the input image, consisting only of pure white lines on an opaque black background. Remove the face, body, and all color, lighting, shading, and other appearance information.
Each line must trace the complete path of a single hair strand. For each hair cluster, draw approximately three evenly spaced lines that clearly describe the cluster’s 3D volume, width, and surface flow. Maintain visible gaps between lines within the same cluster. Ignore lighting and shadows entirely. Keep each line continuous, breaking it only where the strand is genuinely occluded. Every line must follow the hair’s growth direction and align precisely with its position in the input image.
Style: Vector graphic, flat SVG style, computer-generated splines, strictly uniform stroke width, monoline. Binary black-and-white image with hard edges and no anti-aliasing.

\end{quote}

\section{Discussion}

\paragraph{Limitations and future work}
While HairFlow handles a wide range of hairstyles, fine curls in Afro-textured hair and complex strand connectivity remain challenging, as volumetric orientation fields may smooth high-frequency details and do not explicitly represent connectivity. Reconstruction also depends on the input mesh: for example, a ponytail fused with the head can hinder the recovery of separated strands.
Future work could explore spatially localized representations for interactive editing, enabling artists to modify selected hair regions through sketch guidance while preserving the rest of the hairstyle.

\end{document}